\pdfoutput=1

\documentclass[11pt]{article}

\usepackage[]{acl}

\usepackage{times}
\usepackage{latexsym}

\usepackage{graphicx}
\usepackage{verbatim}
\usepackage{physics}
\usepackage{cleveref}
\usepackage{listings}
\usepackage{color}
\usepackage{adjustbox}
\usepackage{hyperref}

\definecolor{dkgreen}{rgb}{0,0.6,0}
\definecolor{gray}{rgb}{0.5,0.5,0.5}
\definecolor{mauve}{rgb}{0.58,0,0.82}

\crefname{section}{\S}{\S\S}
\Crefname{section}{\S}{\S\S}

\usepackage{xcolor}
\definecolor{new-pink}{rgb}{0.89,0.07,0.53}
\definecolor{new-green}{rgb}{0.274,0.502,0.14}
\usepackage[T1]{fontenc}

\usepackage[utf8]{inputenc}

\usepackage{microtype}

\newcommand{\Ek}{\mathcal{E}_k}  

\newcommand{\MainPrompt}[0]{f_{prompt}(x_t, \Ek)}
\newcommand{\InvPrompt}[0]{f'_{prompt}(\Ek)}
\newcommand{\enc}[1]{emb({#1})}
\newcommand{\encoder}{emb}
\newcommand{\truesim}[2]{sim_{F_1}(#1, #2)}
\newcommand{\outline}[1]{}
\newcommand{\nilay}[1]{}
\newcommand{\cm}[1]{}
\newcommand{\gr}[1]{}   

\title{Diverse Retrieval-Augmented In-Context Learning for Dialogue State Tracking}

\author{Brendan King \and Jeffrey Flanigan\\
  University of California, Santa Cruz \\
  \texttt{\{bking2,jmflanig\}@ucsc.edu}}

\begin{document}
\maketitle

\begin{abstract}
There has been significant interest in zero and few-shot learning for dialogue state tracking (DST) due to the high cost of collecting and annotating task-oriented dialogues. Recent work has demonstrated that in-context learning requires very little data and zero parameter updates, and even outperforms trained methods in the few-shot setting \cite{hu_-context_2022}. We propose RefPyDST, which advances the state of the art with three advancements to in-context learning for DST.
First, we formulate DST as a Python programming task, explicitly modeling language coreference as variable reference in Python. Second, since in-context learning depends highly on the context examples, we propose a method to retrieve a diverse set of relevant examples to improve performance. Finally, we introduce a novel re-weighting method during decoding that takes into account probabilities of competing surface forms, and produces a more accurate dialogue state prediction.
We evaluate our approach using MultiWOZ and achieve state-of-the-art multi-domain joint-goal accuracy in zero and few-shot settings.\footnote{Our code: \href{https://github.com/jlab-nlp/RefPyDST}{https://github.com/jlab-nlp/RefPyDST}
}
\end{abstract}

\section{Introduction}
\begin{figure}[h!]
\includegraphics[width=\columnwidth]{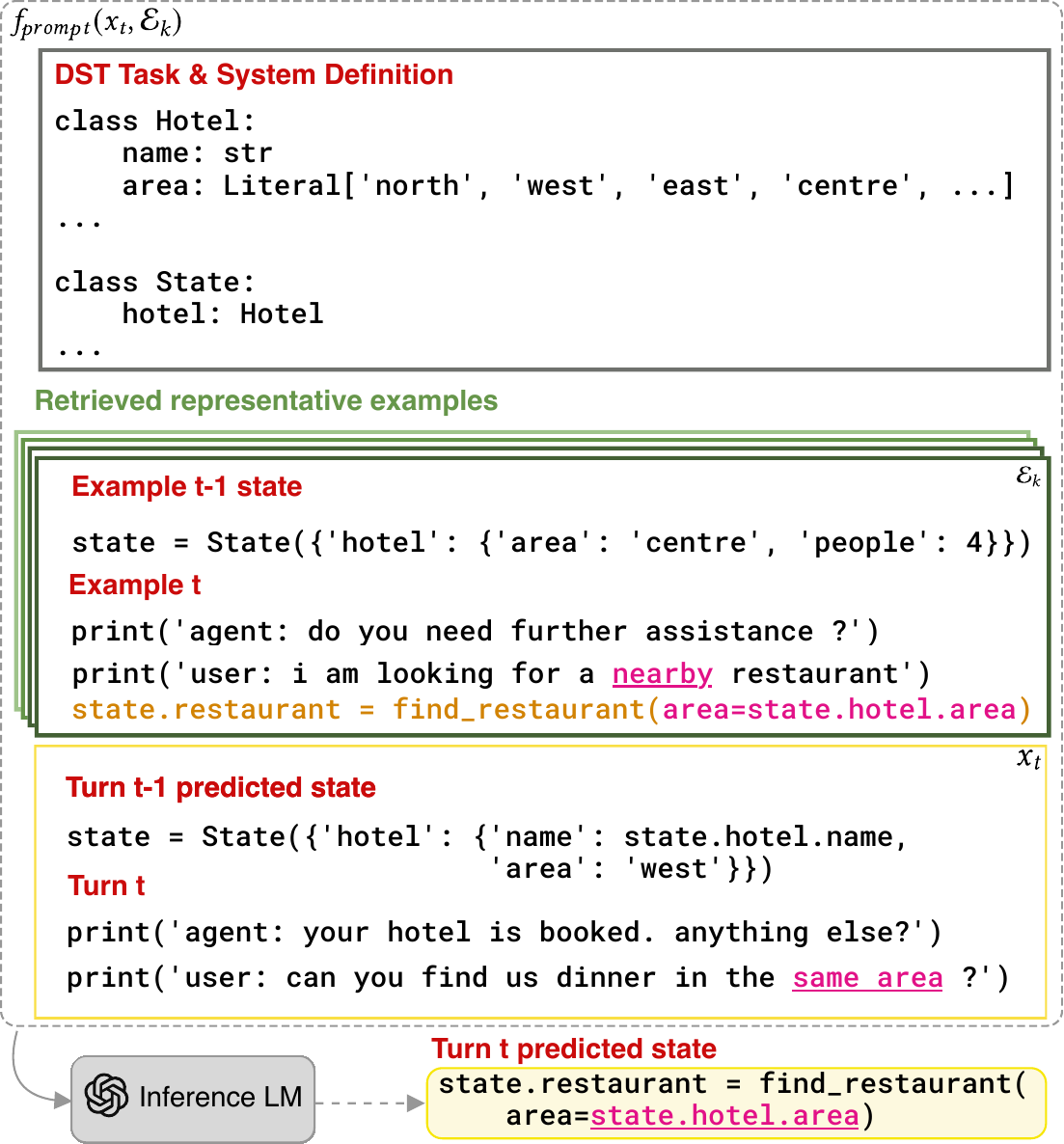}
\caption{Our retrieval-augmented in-context learning approach to DST. We construct a prompt which re-frames DST as a Python programming task conditioned on a system definition and set of retrieved examples $\Ek$ (\color{new-green}{green}\color{black}{}). For each dialogue turn $t$, the goal is to take the current state (\texttt{state}) and turn utterances (\texttt{print(...)}) as `input' and produce a program which \textit{updates} the state with missing values, i.e. $(\text{restaurant-area}, \text{west})$. We represent linguistic coreference explicitly as variable reference (\color{new-pink}{pink}\color{black}{})}
\label{fig:main} 
\centering
\end{figure}

Dialogue state tracking (DST) is an important language understanding task required for supporting task-oriented conversational agents. For each turn in a dialogue, the goal of DST is to extract the intentions and arguments a user communicates into a meaning representation aligned with the capabilities of the system. Often, this can be represented as a set of slot-value pairs, using slots defined in a system schema. For example, if a user asks a hotel booking agent for "a four-star hotel with somewhere to park", the agent could extract the state $\{(\text{hotel-stars}, 4), (\text{hotel-parking}, \text{yes})\}$.

Annotating these turn-level dialogue states is challenging and time-intensive \cite{budzianowski2018large}. Further, as system capabilities evolve over time, the schema and DST requirements change. As such, flexible and data-efficient DST methods are highly valuable.

For these reasons, recent work has explored zero and few-shot methods for DST.  Few-shot methods often fine-tune a pre-trained language model (LM) on DST or a re-framing of the task \citep[e.g.][]{su_multi-task_2021, shin-etal-2022-dialogue, lin-etal-2021-zero}. While these systems are often data efficient, they are inflexible to changing system definitions, requiring re-training as new services are added. To address this, zero-shot methods for domain transfer have been proposed \citep[e.g.][]{wu_transferable_2019, hosseini-asl_simple_2020, gupta_show_2022}, but their performance in new domains can significantly depend on conceptual overlap with training domains \cite{wu_transferable_2019}.

The in-context learning framework (ICL) \cite{brown_language_2020} is particularly appealing in this setting given that it is highly data-efficient and flexible: instead of fine-tuning, ICL methods prompt a fixed LM with templated examples for a task. This approach requires no re-training when adapting to schema changes. In recent work, \citet{hu_-context_2022} find that prompting a language model with examples for DST in a text-to-SQL format can outperform fine-tuned zero and few-shot methods.

In this work, we propose \textbf{RefPyDST}, a retrieval-augmented in-context learning approach to DST for use with language models pre-trained on code, such as OpenAI Codex \cite{chen_evaluating_2021}, by building on recent ICL methods for DST \cite{hu_-context_2022}. Our approach advances the state of the art with three key contributions.

First, we develop a novel in-context prompt that re-frames DST as text-to-python, explicitly modeling slot value coreferents using variables. We provide an overview of this prompt and example of such coreference in \autoref{fig:main}. We demonstrate that this approach significantly improves system performance in the zero and few-shot settings, and particularly improves accuracy on predictions requiring coreference resolution.

Second, we introduce a novel method for diverse supervised example retrieval, which yields a set of in-context examples $\Ek$ that are both individually relevant and collectively representative of the output space, inspired by maximum marginal relevance (MMR) \cite{goldstein-carbonell-1998-summarization}. Our approach significantly improves performance in few-shot settings, overcoming a failure mode in supervised example retrieval in which examples are each similar to an input $x$ but redundant in the outputs they demonstrate.

Third, we propose a novel scoring method $PMI^\beta$ which compensates for surface-form competition among sampled LM completions in constrained generation settings. Inspired by \citet{holtzman-etal-2021-surface}, we re-weigh each completion $y$ by an estimate of its a priori likelihood in the task context. We find this improves system performance in both the zero and few-shot settings.

Together, our contributions address key challenges in DST and in retrieval-augmented ICL generally. Our method produces state-of-the-art results on MultiWOZ 2.1 and 2.4 DST benchmarks across a variety of few-shot settings. Similarly, we obtain a new zero-shot state-of-the-art in the multi-domain setting.

\section{Task Definition}
\label{sec:background}
A task-oriented dialogue consists of turns or paired utterances between a user and an agent which interfaces the user with a programmable system.
At each turn $t$, the purpose of a DST module is to use the dialogue history up to that turn to predict a dialogue state $y_t$, which represents the user's goal and progress in using the system. Let $A_i$ be an agent utterance, $U_i$ be a user utterance, and $C_t = [(A_1,U_1),(A_2,U_2), ... (A_t,U_t)]$\footnote{For user-initiated dialogues, $A_1$ may be omitted} be the dialogue history up to turn $t$. The task is to map the history $C_t$ to a state representation $y_t$.
In this work, we predict dialogue states $y_t$ which can be represented as slot-value pairs:
$$
y_t = \{(s_1, v_1), (s_2, v_2) ... (s_n, v_n)\}
$$
where each slot $s_i$ and the types of values it permits are defined in a system schema. For example, an agent supporting hotel reservations might have a slot `hotel-parking' taking boolean values for constraining search to hotels that include parking.

We can equivalently define this task as predicting \textit{state changes}, as proposed in \citet{hu_-context_2022}. Let $x_t = [y_{t-1}, (A_t, U_t)]$ be a dialogue \textit{context} consisting of the previous dialogue state prediction and utterances for the current turn. Using this turn context $x_t$, we predict a state change:
$$
\Delta y_t = \{+(s_i, v_i) ... -(s_j, v_j) ...\}
$$
where $y_t$ is computed by applying the difference $\Delta y_t$ to $y_{t-1}$. 
This approach has two advantages for few-shot in-context learning. 
First, the turn context $x_t$ requires fewer tokens to represent than the complete history $C_t$, permitting more in-context examples. 
Second, the number of distinct state changes $\Delta y_t$ observed in practice is much smaller than the number of distinct states $y_t$, simplifying the search for relevant examples and the generation problem.

For these reasons, we formulate our DST problem as mapping from the turn context $x_t$ to a state change $\Delta y_t$. For readability, we often use `turn' to refer to this turn context $x_t$, distinguishing it from the history $C_t$ or turn number $t$ using notation. 

\section{Methods}

Given a dialogue turn $t$, our method produces a state change $\Delta y_t$ by (1) retrieving a set of in-context examples $\Ek$, (2) formatting these into a prompt $\MainPrompt$, (3) generating and scoring possible program solutions (LM completions) with OpenAI Codex \cite{chen_evaluating_2021}, (4) executing the program to compute a state change $\Delta y_t$. Given the state change, we compute the complete dialogue state $y_t$ by applying the difference to $y_{t-1}$. We describe our prompting function $\MainPrompt$, in \S ~\ref{sec:methods-prompt}. In \S ~\ref{sec:methods-retrieval}, we describe our method for retrieving a diverse and representative set of examples $\Ek$. Finally, we describe our method for scoring LM completions with a point-wise mutual information estimate in \S ~\ref{sec:methods-scoring}.

\subsection{Prompting with Text-to-Python}
\label{sec:methods-prompt}

We design a novel prompt that re-frames DST as a text-to-Python task, allowing us to explicitly represent coreference phenomena and leverage the unique capabilities of language models pre-trained with code. \autoref{fig:main} provides an overview. Formally, we define a prompting function $\MainPrompt$, which takes a test dialogue turn $x_t$ and a set of $k$ in-context examples $\Ek = \{(x_1, \Delta y_{1}), ...(x_k, \Delta y_{k})\}$ and produces a string representing the program synthesis task.

Our prompt (\autoref{fig:main}) starts with a task definition represented as a set of Python classes corresponding to each DST domain. Each informable slot is an attribute in the appropriate class. Type hints are used to label categorical slots with their values and non-categorical slots with the most appropriate type. The dialogue state is also represented as an object which can be manipulated, having an attribute per-domain.

We represent instances of our programming synthesis task with in-context python examples. Each in-context example $([y_{t-1}, A_t, U_t], \Delta y_t)$ is represented as follows: the previous dialogue state $y_{t-1}$ is represented as a dictionary, mapping slot names to values. 
Non-categorical values such as names are de-lexicalized by replacing their string value with a variable referencing their existing value in the state.
Solutions to the programming task are represented as function calls that manipulate the dialogue state. One of the key benefits of our formulation of the DST task as python is explicit representation of coreference phenomena. For example, the solution corresponding to a user input ``find me a restaurant in the same area as my hotel" would be \texttt{state.restaurant = find\_restaurant(area = state.hotel.area)}, explicitly modeling the resolution of the linguistic coreference.

\subsection{Retrieving Diverse Relevant Examples}

\label{sec:methods-retrieval}

We propose a method for in-context example selection that produces an example set $\Ek$ that is both relevant to a test turn $x_t$ and diverse, representing the relevant portions of the output space. We first learn an embedding space in which similar state changes have high cosine similarity with one another (\S \ref{sec:methods-approx-relevance}), following \cite{hu_-context_2022}. Using this, we propose a novel method for decoding $\Ek$ such that examples are similar to $x_t$ but dissimilar to each other (\S \ref{sec:methods-diverse-decoding}).

\subsubsection{Retriever Training}
\label{sec:methods-approx-relevance}
We fine-tune an embedding model to approximate the true similarity between two turn contexts $x_i, x_j$ with the \textit{cosine similarity} between their encoded representations, following prior works \cite{hu_-context_2022, rubin_learning_2021}. Let $D_{train}$ be a set of dialogue turns serving as training data for an example retriever and selection pool at inference time. As described in \S \ref{sec:background}, each example $e_i \in D_{train}$ is a context state-change pair $e_i = (x_i, \Delta y_i)$. A single example $e_i$ is shown in the green box in \autoref{fig:main}.

We encode an example or query turn context $x = [y_{t-1}, (A_t, U_t)]$ by concatenating each element of the turn context and passing the result through an embedding model\footnote{We use all-mpnet-base-v2 \cite{song_mpnet_2020}, available in sentence-transformers \cite{reimers-2019-sentence-bert}} $\encoder$. For two example turn contexts $x_i, x_j$, the cosine similarity between their embeddings $cos(\enc{x_i}, \enc{x_j})$ approximates their relevance to each other. At inference time, we can embed a test turn $x_t$ and retrieve highly similar examples with nearest neighbors search.

We fine-tune our embedding model with a supervised contrastive loss, such that high cosine similarity of representations correlates with high similarity between dialogue state changes, following the procedure in \citet{hu_-context_2022}. For our learning objective, we assume a metric that gives the \textit{true} similarity between two dialogue state changes for a pair of turns $sim_{F_1}$, which we define below. For each dialogue turn in the training set, we use $sim_{F_1}$ to define positive and (hard) negative examples as the top and bottom 5\% of the current nearest 200 examples, respectively. We train each retriever for 15 epochs using the hyperparameters detailed in \autoref{sec:appendix-impl-details}.

We define the ground-truth similarity $sim_{F_1}$ between two dialogue state changes as follows. Let $\Delta y^a = \{(s_1^a, v_1^a)... (s_m^a, v_m^a)\}$ and $\Delta y^b = \{(s_1^b, v_1^b)... (s_n^b, v_n^b)\}$ be two dialogue state changes. For any slot value $v_i$ exhibiting coreference to another slot $s_j$, we replace $v_i$ with $s_j$. For example, the state change corresponding to a turn "I need a taxi to my hotel" would become $\{(\text{taxi-destination}, \text{hotel-name})\}$, regardless of the particular hotel name value. 
We then compute true state similarity using the average between the $F_1$ score comparing updated slots and the $F_1$ score comparing updated slot-value pairs, as proposed in \citet{hu_-context_2022}:
\begin{align*}
    \truesim{\Delta y^a}{\Delta y^b} = \frac{1}{2}F_1(\{s_1^a, ...\}, \{s_1^b, ...\}) + \\ \frac{1}{2}F_1(\{(s_1^a, v_1^a), ...\}, \{(s_1^b, v_1^b), ...\})
\end{align*}

\subsubsection{Decoding Diverse Examples}
\label{sec:methods-diverse-decoding}

We propose an adaptation of MMR which uses our learned embedding model $\encoder$ to produce a diverse set of examples $\Ek$ that maximizes similarity to $x_t$ and minimizes similarity between examples in $\Ek$. Particularly for encoders that are fine-tuned to approximate output similarity, this yields a set of examples that is more representative of the output space than simply selecting the nearest $k$, which may all have the same label. Formally, we define the ideal set of in-context examples $\mathcal{E}^*_k$ for an input $x_t$ to be the $k$ examples satisfying:
\begin{align*}
    \mathcal{E}^*_k  = \mathop{argmax}_{\mathcal{E}_k \subset \mathcal{D}_{train}} \sum_{x_i \in \Ek} cos(\enc{x_t}, \enc{x_i}) \\ - \alpha \sum_{x_i, x_j \in \Ek}  cos(\enc{x_i}, \enc{x_j})
\end{align*}
where the hyperparameter $\alpha$ is a dissimilarity factor and $\alpha=0$ corresponds to typical nearest-$k$ example selection. We greedily approximate $\mathcal{E}^*_k$ by iteratively selecting the example which maximizes the equation at each step.
For more efficient decoding of $\Ek$ with large selection pools, we limit the considered examples to the nearest $N$ such that $|D_{train}| >> N >> k$. For example in one run in the 5\% MultiWOZ few-shot setting, $|D_{train}| = 2754$, $N=100$, and $k=10$.

\subsection{Decoding with Point-wise Mutual Information}
\label{sec:methods-scoring}

 \begin{figure*}[t]
     \centering
     \includegraphics[width=0.9\textwidth]{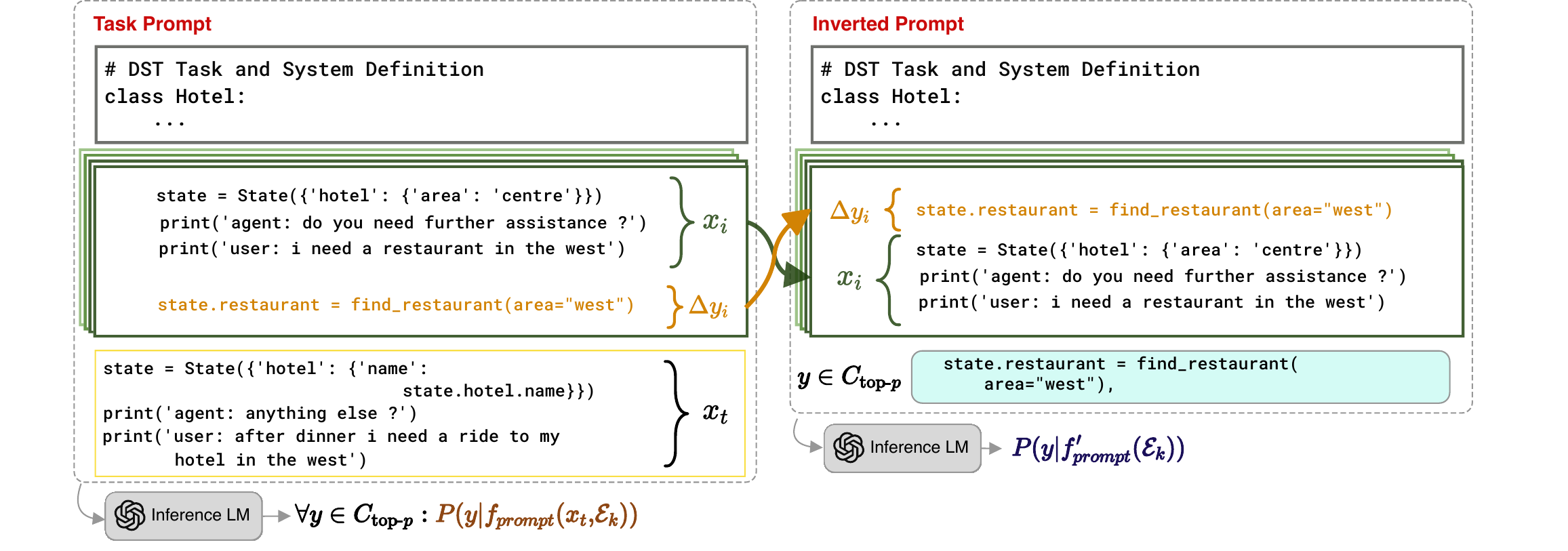}
     \caption{An overview of our method (\S \ref{sec:methods-scoring}) for scoring completions $y$ from Codex with $PMI^\beta$, which re-weighs using an estimate of the a priori likelihood of $y$ in the context of the task. On the left, is our primary text-to-Python prompt $\MainPrompt$ (\autoref{sec:methods-prompt}). We use nucleus sampling to generate a set of reasonable candidates $C_{\text{top-}p}$ and their probabilities. On the right is an inverted prompt with state changes preceding their inputs, allowing us to produce an in-context estimate of the probability of $y$ not conditioned on $x$}
     \label{fig:prompts}
 \end{figure*}

We introduce a new rescoring function, $PMI^\beta$, to mitigate surface form competition when generating from language models, that we use for making predictions in our setting.  $PMI^\beta$ is an extension of $PMI_{DC}$, which was proposed in \citet{holtzman-etal-2021-surface} for mitigating surface form competition in the classification setting.
We first describe surface form competition and $PMI_{DC}$ (\S \ref{sec:surface-form-competition}), and then describe $PMI^\beta$, an adaptation of this method to the constrained generative setting with in-context examples (\S \ref{sec:methods-pmi-beta}).

\subsubsection{Surface-form Competition}
\label{sec:surface-form-competition}
Conditioned on a prompt, a language model assigns a likelihood to all completing strings, from which we can sample. While string likelihoods can be used as a proxy for output class or structure likelihoods, these are not the same. For example, in our DST formulation, many strings can correspond to the same state change $\Delta y_t$, or may not correspond to a valid state change at all. As such, \citet{holtzman-etal-2021-surface} argue string likelihoods can be unreliable for scoring the best among a fixed set of choices which may each contain numerous surface forms in $V^*$. To compensate for this, they propose scoring with Domain Conditional Point-wise Mutual Information ($PMI_{DC} = \frac{P(y|x, domain)}{P(y|domain)}$). This re-weighs choices by a priori likelihood of their string form in the task context $P(y|domain)$.

\subsubsection{Scoring with $PMI^\beta$}
\label{sec:methods-pmi-beta}
To mitigate surface-form competition, we propose $PMI^\beta$: a prompt conditional pointwise mutual information scoring method that adapts $PMI_{DC}$ to our constrained generative setting with in-context examples. Doing so requires overcoming two key challenges. First, our choices to score amongst are not practically enumerable. Second, the task context we condition on is partly defined by our choice of in-context examples $\Ek$. We overcome these by first generating a small set of plausible completions $\mathcal{C}$ and their likelihoods according to a language model. Then, we re-weigh these likelihoods according to an estimate of their a priori likelihood conditioned on only the task context and selected examples $\mathcal{E}_k$:
\begin{equation}
\label{eq:prompt-pmi}
PMI^\beta(x;y|\mathcal{E}_k) = \frac{P(y|\MainPrompt))}{P(y|\InvPrompt)^\beta}
\end{equation}
where $f'_{prompt}$ is a prompt designed for estimating $P(y|\mathcal{E}_k)$ without conditioning on $x_t$, described below, and $\beta$ is a hyperparameter for adjusting the impact of re-weighing by a priori likelihood.\footnote{While only $\beta=1$ corresponds neatly to a point-wise mutual information estimate $pmi(x_t;y)$, we find $0 < \beta < 1$ to be more effective in practice. Prior work in terminology extraction has also proposed scaling PMI estimates, though in a different context \cite{Daille1994ApprocheMP}} 

To generate the candidate completions $\mathcal{C}$, we sample a set of plausible candidates using nucleus sampling \cite{holtzman_curious_2020}.

While one could simply use the language model to compute $P(y)$ directly, such unconditional estimates tend to vary wildly. Following \citet{holtzman-etal-2021-surface}, we instead estimate the probability of the completion in context, but further account for the use of in-context examples. To do this, we construct an additional prompt which contains the same problem definition, but reverses the order outputs and inputs. Using this, we can estimate the probability of a completion $y$ in the context of our task and examples without $x_t$, illustrated in \autoref{fig:prompts}. Finally, we select the completion $\hat{y}$ which maximizes Eq.~\ref{eq:prompt-pmi}, and parse it to a dialogue state change $\Delta y_t$:

$$
\hat{y} = \mathop{argmax}_{y \in \mathcal{C}} PMI^\beta(x;y|\mathcal{E}_k)
$$

We choose a minimum a priori likelihood of between $10^{-7}$ and $10^{-5}$, as estimates for $P(y|f'_{prompt}(\mathcal{E}_k))$ can be very low, particularly when rare slot values implied by $x_t$ are not present in any example. When constructing our candidate set $\mathcal{C}$, we choose the five most likely sampled completions under the original prompt. Finally, we canonicalize each completion $y$ when computing $P(y|f'_{prompt}(\mathcal{E}_k))$ by first parsing it to a dialogue state change, and then re-writing it as a string in the form as if it were an example in $\mathcal{E}_k$. In effect, this normalizes mis-spellings and enforces the expected order of keyword arguments in the update string, further controlling for high variance in our estimates.

\section{Experiments}

We describe our zero and few-shot experimental setups, evaluation, and baselines. Hyperparameter and implementation details can be found in \autoref{sec:appendix-impl-details}.

\begin{table*}[h!]
\centering
\resizebox{\textwidth}{!}{
\begin{tabular}{lrrrr|rrrr}
\hline
 &  \multicolumn{4}{c}{MultiWOZ 2.1} & \multicolumn{4}{c}{MultiWOZ 2.4}  \\
Model & 1\% & 5\% & 10\% & 100\% & 1\% & 5\% & 10\% & 100\% \\
\hline
TRADE \cite{wu_transferable_2019} & 12.6 & 31.2 & 36.2 & 46.0 & - & - & - & 55.1 \\
DiSTRICT \cite{venkateswaran_district_2022} & 13.4 & 41.3 & 49.7 & 56.1 & - & - & - & - \\
DS2 \cite{shin-etal-2022-dialogue}  & 33.8 & 44.2 & 45.4 & 52.3 & 36.8 & 49.9 & 51.1 & 57.9 \\
\hline
IC-DST Codex \cite{hu_-context_2022} & 43.1 & 47.1 & 48.7 & 50.7 & 48.4 & 55.4 & 56.9 & 62.4 \\

RefPyDST (ours) & \textbf{47.3} & \textbf{49.6} & \textbf{50.8} & 52.0 & \textbf{55.2} & \textbf{62.3} & \textbf{62.5} & 65.2 \\
\end{tabular}}
\caption{Multi-domain JGA evaluated on MultiWOZ 2.1 \& 2.4 using samples from  1\%, 5\%, 10\%, and 100\% of the training set. Average of three runs is reported. Our method achieves state-of-the-art (\textbf{bolded}) for both dataset versions in the 1\%, 5\%, and 10\% few-shot settings. Our method also out-performs all few-shot baselines which report results in the 100\% setting on MultiWOZ 2.4. Line distinguishes fine-tuned from in-context learning methods.}
\label{tab:few-shot-results}
\end{table*}

\subsection{Experimental Settings}

We conduct zero and few-shot DST experiments on the MultiWOZ dataset \cite{budzianowski2018large}, containing over ten thousand multi-domain task-oriented dialogues crowd-sourced in a wizard-of-oz setup. There are five domains in the validation/test sets and a total of thirty informable slots. We evaluate on the newest MultiWOZ 2.4 \cite{ye_multiwoz_2022}. For comparison with prior work, we also report on MultiWOZ 2.1 \cite{eric_multiwoz_2020}.

We evaluate performance with standard joint-goal accuracy (JGA) for all of our experiments. For a turn $x_t$, a dialogue state prediction $\hat{y}_t$ is considered correct only if all slot names and values exactly match the ground-truth state $y_t$.

For the few-shot setting, following \cite{wu_improving_2020}, we sample 1\%, 5\%, or 10\% of the dialogues from the training set to serve as a training set $D_{train}$ for each experiment. We fine-tune our retriever using $D_{train}$ and select in-context examples from it. We conduct three independent runs for each sample size and report the average JGA across runs. We also perform a single run in the full setting, using 100\% of the training data.

For the zero-shot setting, there are no labeled examples to select from, but a single formatting example is used for all inference turns, as in \cite{wang-etal-2022-benchmarking, hu_-context_2022}. We consider two evaluation settings. The first is the typical assessment on all test set dialogues, as in few-shot and complete training regimes, which we will refer to as the standard MultiWOZ benchmark. These results allow comparison to few-shot and full-data results, as well as other methods which use zero supervised dialogues in training. We also report results on the MultiWOZ `leave-one-out' benchmark for zero-shot transfer methods \cite{wu_transferable_2019}, reporting JGA considering only slots in each individual domain, as well as the average of these five single-domain results.

We compare to a number of prior state-of-the-art zero-shot and few-shot DST methods as baselines. These include DST specific architectures \cite{wu_transferable_2019}, various fine-tuning methods \cite{gupta_show_2022, shin_few-shot_2022, venkateswaran_district_2022}, and a strong ICL baseline \cite{hu_-context_2022}.

\begin{table*}[h!]
\centering
\begin{tabular}{lrrrrr|r}
\hline
& attraction & hotel & restaurant & taxi & train & \textbf{Avg.} \\
\hline
\multicolumn{7}{c}{\textbf{MultiWOZ 2.1}} \\
\hline
TRADE \cite{wu_transferable_2019} $\dag$ & 20.1 & 14.2 & 12.6 & 59.2 & 22.4 & 25.7 \\
TransferQA \cite{lin-etal-2021-zero} $\dag$ & 31.3 & 22.7 & 26.3 & 61.9 & 36.7 & 35.8 \\
DiSTRICT \cite{venkateswaran_district_2022} $\dag$ & 33.4 & 22.4 & 24.0 & 66.6 & 47.7 & 38.8  \\
D3ST \cite{zhao_description-driven_2022} $\dag$ & 56.4 & 21.8 & 38.2 & 78.4 & 38.7 & 46.7 \\
SDT-seq \cite{gupta_show_2022} $\dag$ & \textbf{74.4} & 33.9 & \textbf{72.0} & \textbf{86.4} & 62.9 & \textbf{65.9}  \\
\hline
 IC-DST \cite{hu_-context_2022} & 60.0 & 46.7 & 57.3 & 71.4 & 49.4 & 57.0 \\
 RefPyDST (ours) & 70.9 & \textbf{51.2} & 65.6 & 67.1 & \textbf{69.2} & 64.7 \\
\hline
\multicolumn{7}{c}{\textbf{MultiWOZ 2.4}} \\
\hline
 IC-DST Codex \cite{hu_-context_2022} & 62.1 & 53.2 & 54.9 & \textbf{71.9} & 51.4 & 58.7 \\
 RefPyDST (ours) & \textbf{74.5} & \textbf{56.6} & \textbf{68.2} & 68.5 & \textbf{76.1} & \textbf{68.8} \\
\end{tabular}
\caption{Zero-shot joint-goal accuracy (JGA) for each domain in MultiWOZ 2.1 \& 2.4 in the leave-one-out set up. We report results on each held-out domain and the average held-out domain performance (Avg.) Domain transfer methods (marked with \dag) learn from dialogues in the other four domains and are tested on the held-out domain. Unlike domain transfer methods, IC-DST and our method do not use any DST data. Following prior work, we evaluate only dialogues and slots in the held-out domain. 
For full evaluation of all dialogues in the zero-shot setup, see Table~\ref{tab:zero-shot-multi}.}
\label{tab:zero-shot-leave-one-out}
\end{table*}
\section{Results}

\paragraph{Few-shot DST on MultiWOZ} We present few-shot and full-shot dialogue state tracking results on MultiWOZ 2.1 \& 2.4 in \autoref{tab:few-shot-results}. We find that our method achieves state-of-the-art in the 1\%, 5\%, and 10\% few-shot settings for both MultiWOZ 2.1 \& 2.4, outperforming all fine-tuned methods as well as other in-context learning methods. While all methods considered improve with additional data, our method is remarkably data efficient: RefPyDST achieves 95\% of its full-shot performance using only 5\% of the training data, on average. In comparison, using 5\% of the training data with IC-DST Codex only achieves 89\% of its full-shot performance.

\begin{table}[]
    \centering
    \begin{tabular}{l|r}
         \hline
         \multicolumn{2}{c}{\textbf{MultiWOZ 2.4}}\\
         \hline
         IC-DST Codex \cite{hu_-context_2022} & 35.3 \\ 
         RefPyDST (ours) & \textbf{47.9} \\
    \end{tabular}
    \caption{Zero-shot (zero DST training data) multi-domain JGA evaluated on MultiWOZ 2.4. Our method achieves state-of-the-art for this setting. Comparisons with zero-shot transfer methods, which train on subsets of the MultiWOZ dataset, can be found in \autoref{tab:zero-shot-leave-one-out}.}
    \label{tab:zero-shot-multi}
\end{table}

\paragraph{Zero-shot DST on MultiWOZ} We present zero-shot multi-domain results on MultiWOZ 2.4 in \autoref{tab:zero-shot-multi}. We find our method outperforms all zero-shot methods, achieving a 12.4\% increase in multi-domain JGA over IC-DST Codex, our strongest performing baseline. Comparisons are limited to methods that use zero training data, as opposed to transfer methods that train on some MultiWOZ domains and evaluate on others. 

For comparison with domain transfer methods, we present zero-shot results on the leave-one-out benchmark for MultiWOZ 2.1 \& 2.4 in \autoref{tab:zero-shot-leave-one-out}. 
Following prior work, we evaluate only dialogues and slots in the held-out domain.\footnote{Prior work on the leave-one-out
setting evaluates using the following method: (1) filter to dialogues which \textit{contain} the held out domain (this can include dialogues in multiple domains) and (2) only check slots in that domain when computing JGA. \cite{wu_transferable_2019}}
Evaluating average performance in this setting, we find our method outperforms all methods except for the current state-of-the-art transfer method, SDT-seq. Their method outperforms ours by 1.5\% on each held-out domain on average. However, transfer methods such as SDT-seq require significant out-of-domain DST training data, while ours requires none. Despite this training data disadvantage, our approach outperforms all other zero-shot transfer methods.

\section{Analysis \& Ablations}
\label{sec:results-ablations}

In this section, we further analyze the performance characteristics of our method. 

\paragraph{Ablations}

\begin{table}[h]
    \centering
    \begin{tabular}{l|r}
        \multicolumn{2}{c}{\textbf{Few-Shot (5\%)}} \\
        \hline
        IC-DST (baseline) & 52.4 \\
        \hline
        RefPyDST -- Python & 54.8 \\
        RefPyDST -- diverse & 54.6 \\
        RefPyDST -- $PMI^\beta$ & 56.1 \\
        \hline
        RefPyDST (full) & \textbf{57.9}  \\
        \multicolumn{2}{c}{} \\
        \multicolumn{2}{c}{\textbf{Zero-Shot}} \\
        \hline
        IC-DST (baseline) & 43.0 \\
        \hline
        RefPyDST -- Python & 40.7 \\
        RefPyDST -- $PMI^\beta$ & 46.0 \\
        \hline
        RefPyDST (full) & \textbf{46.7}  \\

    \end{tabular}
    \caption{MultiWOZ joint-goal accuracy in the few-shot (5\%) and zero-shot settings, leaving out individual components of our method. We evaluate on a 20\% sample of the development set (200 dialogues). For few-shot, we average over three runs, each with independently sampled $D_{train}$. For ablating the removal of our Python prompt, we use the Text-to-SQL format from \cite{hu_-context_2022} as a baseline. The alternatives to our diverse retrieval approach and $PMI^\beta$ scoring are top-$k$ retrieval and greedy decoding, respectively}
    \label{tab:ablation}
\end{table}

In order to assess how each part of our method contributes to performance, we conduct a leave-one-out ablation, as well as reporting the performance of using only our prompting method. Each ablation is conducted using a 20\% sample of the development data in the MultiWOZ 2.4 dataset (200 dialogues), sampled independently of the set used to tune hyperparameters. We present results in \autoref{tab:ablation} for the zero and 5\% few-shot setting. In the few-shot setting, we find leaving out our diverse retrieval to be most impactful.

\paragraph{Does using Python improve coreference resolution?}

Since our Python prompting method explicitly models coreference through variable reference, we analyzed how our system performed on state predictions requiring coreference resolution. Using coreference annotations released with the 2.3 version of the MultiWOZ dataset \cite{han_multiwoz_2021}, we evaluate accuracy on slot values which require coreference to resolve. Our results are presented in \autoref{tab:coreference-results}. Overall, our full model improves upon the baseline for coreference. Removing Python greatly reduces our model's performance, demonstrating the benefit of modeling coreference as Python variable reference.

\begin{table}[h]
    \centering
    
    \begin{tabular}{l|r|r}
        \hline
        \textbf{Model} & 0\% & 5\% \\
        \hline 
         IC-DST (baseline) & $67.7$ & $78.9^*$ \\ \hline
         RefPyDST (prompt only) & $77.1^*$ & $77.9^*$  \\
         RefPyDST  -- Python & $62.9$ & $73.0$  \\
         RefPyDST (full) & $76.8^*$ & $81.8$ \\
    \end{tabular}
    \caption{Accuracy on slot value predictions which require coreference resolution for the zero-shot (0\%) and few-shot (5\%). For a given setting (column), $^*$ indicates the difference is not statistically significant. All other differences in a column are significant to $p<0.02$} 
    \label{tab:coreference-results}
\end{table}

\paragraph{Does our retrieval method improve demonstrated label diversity?}

We investigate to what degree our diverse decoding procedure increases diversity in the distribution of demonstrated labels for a given input. To approximate a label, we define $S(e_i)$ as the distinct combination of \textit{slot names} in the output for an in-context example $e_i = (x_i, \Delta y_i)$, ignoring assigned values.

First, we simply count the average number of distinct combinations of slot names in $\Ek$, shown in upper half of \autoref{tab:diversity-analysis}. For each $x_t$, we retrieve a set of in-context examples $\Ek$. We count the number of distinct slot combinations across each $e_i \in \Ek$, and report the development set average. A value of 1 indicates the retriever is fully redundant: all $k$ examples demonstrate the same combination of slots, while a value of $k$ indicates every example in $\Ek$ is unique.

Second, we consider the entropy of slot combinations present in $\Ek$, shown in the lower half of \autoref{tab:diversity-analysis}. For each $x_t$, we again compute $S(e_i)$ for each retrieved example in $\Ek$. We then compute the specific conditional entropy $H(S|X = x_t)$, estimating the probability of each slot combination $p(S|x_t)$ using its frequency in $\Ek$. We report the development set average or conditional entropy $H(S|X)$. $H(S|X = x_t) = 0$ indicates a fully redundant retriever that retrieves the same set of slots for all examples, and a uniform distribution of slot combinations yields $H(S|X = x_t) = log_2(k)$.\footnote{While this is true of a uniform distribution over demonstrated slot combinations, we find uniformly sampling from $D_{train}$ yields an entropy of $\sim 2.6$, as the distribution of labels in the training data is not uniform.} 

\begin{table}[]
    \centering
    \begin{tabular}{l | r r r r }
          \multicolumn{5}{c}{\textbf{Number of Distinct $S$ in $\Ek$}} \\
          \hline
          \multicolumn{1}{c}{} & \multicolumn{1}{c}{1\%} & \multicolumn{1}{c}{5\%} & \multicolumn{1}{c}{10\%} & \multicolumn{1}{c}{100\%} \\
         \hline
         random & 7.1 & 7.2 & 7.2 & 7.3 \\
         top-k  & 3.4 & 2.2 & 1.8 & 1.5  \\
         \hline
         diverse ($\alpha=.2$) & \underline{5.3} & \underline{4.1} & 3.3 & 2.2 \\ 
         diverse ($\alpha=.3$) & 5.7 & 4.5 & \underline{3.5} & 2.3 \\
         diverse ($\alpha=.5$) & 7.5 & 5.7 & 4.8 & \underline{2.8} \\
          \multicolumn{5}{c}{} \\
          \multicolumn{5}{c}{\textbf{Entropy $H(S|X)$}} \\
          \hline
          \multicolumn{1}{c}{} & \multicolumn{1}{c}{1\%} & \multicolumn{1}{c}{5\%} & \multicolumn{1}{c}{10\%} & \multicolumn{1}{c}{100\%} \\
         \hline
         random & 2.6 & 2.6 & 2.6 & 2.6 \\
         top-k  & 1.2 & 0.63 & 0.47 & 0.30 \\
         \hline
         diverse ($\alpha=.2$) & \underline{1.8} & \underline{1.5} & 1.1 & 0.64 \\ 
         diverse ($\alpha=.3$) & 1.9 & 1.6 & \underline{1.2} & 0.68 \\
         diverse ($\alpha=.5$) & 2.7 & 2.0 & 1.7 & \underline{0.93} \\
    \end{tabular}
    \caption{We analyze the \textit{outputs} demonstrated in $\Ek$ for different in-context example retrieval methods. Above, we show the average number of distinct slot combinations demonstrated in $\Ek$. Below, we show the conditional entropy $H(S|X)$ of the distribution of slot combinations in $\Ek$. We underline the values corresponding to methods used in our final models}
    \label{tab:diversity-analysis}
\end{table}

We find our retrieval methods increase the diversity of in-context examples across all settings. For a given training set size, we see that diverse decoding increases the number of distinct `labels', measured by $S(e_i)$, as well as the entropy $H(S|X)$. Still, selected examples are not random, as we can see when comparing $H(S|X)$ to a random retriever which uniformly samples from $D_{train}$.\footnote{In Appendix \ref{sec:appendix-random-retrieval}, we also compare few-shot task performance for our retrieval method against random retrieval} Finally, we see that as the size of the training set increases, the diversity in exemplified labels for a given choice of $\alpha$ \textit{decreases}. Increasing training data leads to a higher density of each slot combination, requiring more aggressive discounting to achieve the same diversity in $\Ek$. As such, we increase $\alpha$ with training set size, using $\alpha=0.2$ for 1\% and 5\% settings and $\alpha=0.3$ \& $\alpha=0.5$ for 10\% and 100\% settings, respectively.

\section{Related Work}

\paragraph{Dialogue State Tracking}
There has been a recent increase in work on the zero and few-shot DST systems. Many approaches fine-tune a pretrained language model by re-framing DST as some form of text-to-text or auto-regressive language modeling task \cite{wu_improving_2020, peng_soloist_2021, hosseini-asl_simple_2020, su_multi-task_2021, shin-etal-2022-dialogue, lin_leveraging_2021, gupta_show_2022, li-etal-2021-zero, xie_unifiedskg_2022}. Many of these methods often exhibit zero-shot transfer capabilities \cite{wu_transferable_2019, gupta_show_2022, li-etal-2021-zero, hosseini-asl_simple_2020}. However, these approaches still require re-training when a domain is added or changed, and zero-shot transfer performance is dependent on the relatedness of the new domain to existing ones.

Some recent works instead model DST as an in-context learning problem \cite{hu_-context_2022, xie_unifiedskg_2022, madotto_few-shot_2021}, bypassing the need for re-training when system definitions change.
In particular, we build on the work of \citet{hu_-context_2022}, which models DST by predicting dialogue state \textit{changes} at each turn, relying on only a state summary and agent/user turn utterances for inference. 
Their work models DST as a text-to-SQL problem, whereas we model it as a Python programming problem with novel methods for selecting in-context examples and scoring language model completions. 

\paragraph{In-Context Learning}
Some recent works explore the properties of effective in-context examples. In classification settings, \citet{gao_making_2021} find random examples can significantly limit performance, and propose using a pre-trained embedding model to find examples semantically close to $x$, retrieving one per class. 
Other works investigate the role of examples in ICL performance in detail, finding that ICL methods perform best when example inputs and test inputs are as close in distribution as possible, and when the distribution of exemplified labels closely matches the target distribution \cite{min_rethinking_2022, liu-etal-2022-makes}.

Paralleling this, a number of works across NLP tasks propose methods for retrieving relevant in-context examples. \citet{pasupat-etal-2021-controllable} use an unsupervised embedding model to embed a test input $x$ and all available examples, retrieving the $k$ with highest embedding cosine similarity. Other works use a similar dense retriever but in an embedding space learned with supervision. \citet{rubin_learning_2021} fine-tune an example retriever with contrastive learning in which positive examples maximize $p_{LM}(y|x, e_i)$. \citet{hu_-context_2022} propose a contrastive learning objective specific to DST, fine-tuning an embedding model to embed turns with similar state changes in proximity to each other. Rather than use a separate retrieval module, \citet{shin_few-shot_2022} use the LM itself to select examples which are most likely when conditioned on $x$. Given a test input $x$, each of these works scores the relevance of an individual example $e_i$ to a test input $x$ and then selects the $k$ most relevant ones to include in a prompt. In most cases, this yields a set of examples $\Ek$ which are meaningfully similar to $x$. However, considering examples individually does not necessarily lead to adequate exemplification of the output space. In supervised settings that learn a relevance metric which approximates output similarity, this can lead to degenerate examples sets $\Ek$ which all exemplify the same output. In contrast to this, we propose a novel method for using this score to construct $\Ek$ with examples that are relevant to $x$ while being distinct from each other.

In concurrent work to our own, \citet{ye_complementary_2022} propose a method for decoding diverse examples of explanations from a retriever for use in reasoning problems, also based on maximum-marginal-relevance (MMR) \cite{goldstein-carbonell-1998-summarization}. Their work uses unsupervised measures of similarity between explanations, where ours uses a supervised retriever which approximates similarity of outputs. Thus, diversity in our example sets correlates to diversity in exemplified outputs. In another concurrent work to our own \cite{levy_diverse_2022} propose a method for diverse example selection in a semantic parsing task, using the outputs of selected examples to incrementally cover more structures in $\Ek$. 

For tasks which can be re-framed as program synthesis, a number of works have also developed ICL methods for use with LMs pre-trained on code such as Codex and Codegen \cite{chen_evaluating_2021, nijkamp_codegen_2022}. \citet{shin_few-shot_2022} use ICL with Codex to generate Lisp-like programs in a dialogue semantic parsing task. \citet{Rajkumar2022EvaluatingTT} evaluate such models capabilities in Text-to-SQL problems, and \citet{hu_-context_2022} use a Text-to-SQL framing to use Codex for DST. Instead of SQL queries, we generate Python programs, allowing for intuitive modeling of phenomena like coreference.

Finally, recent works have considered adjusting how completion strings are scored with an LM. \citet{brown_language_2020} normalize log-likelihoods by length before scoring completions. \citet{zhao_calibrate_2021} re-weigh LM probabilities by learning an affine transformation that yields uniform scores given `content-free inputs'. \citet{holtzman-etal-2021-surface} propose $PMI_{DC}$, a method for re-scoring completions using pointwise mutual information (pmi), which we adapt to our constrained generative setting.

\section{Conclusion}

We propose RefPyDST, an in-context learning method for DST. Our contributions address key challenges in DST and in retrieval-augmented ICL, producing state-of-the-art results on MultiWOZ DST benchmarks for few-shot and zero-shot setups. Future work could apply methods developed here to other in-context learning problems. 

\section{Limitations}
While in-context learning methods for DST are promising in their data efficiency and flexibility to new domains, they typically require very large models to perform effectively. At 175 billion parameters, OpenAI Codex \cite{chen_evaluating_2021} is much larger than some of the fine-tuned approaches to DST, though with better performance and ability to adapt to new domains without re-training. Despite our advances, there are still significant errors when applying ICL for DST. As such, ICL may not necessarily be relied on in safety-critical settings.

\section*{Acknowledgements}
We thank Geetanjali Rakshit, Nilay Patel, Changmao Li, Chris Toukmaji, Rongwen Zhao, and other JLab members for insightful feedback on preliminary drafts of this work, and thank the anonymous reviewers and area chairs for their detailed and helpful feedback. The authors were supported in part by the NSF National
AI Institute for Student-AI Teaming (iSAT) under
grant DRL 2019805. The opinions expressed are
those of the authors and do not represent views
of the NSF. We are thankful for the computing resources provided by the Pacific Research Platform's Nautilus cluster, supported by the National Science Foundation under Award Numbers CNS-1730158, ACI-1540112, ACI1541349, OAC-1826967, the University of California Office of the President, and the University of California San Diego’s California Institute for Telecommunications and Information Technology/Qualcomm Institute.
\bibliography{anthology,custom, BetterFewShotTOD}
\bibliographystyle{acl_natbib}

\appendix

\section{Dialogue State Normalization}
\label{sec:appendix-normalization}

Real world task oriented dialogue systems can interface users with thousands or more entities, such as restaurants or hotels in MultiWOZ. Since reasoning directly over all such entities is intractable, dialogue understanding modules often first predict a surface form (e.g. a restaurant name mentioned by a user) which another module links to a canonical form (e.g. that restaurants name in a database). While dialogue state trackers evaluated on MultiWOZ do not need to interact with a database, handling of typos and unexpected surface forms is important for a realistic assessment of system performance, since predictions for a slot are evaluated on exact string match.
As such, most research systems including the baselines in this paper use rule-based functions to fix typos and unexpected surface forms. We propose a robust rule-based method for effective linking of surface forms to canonical forms described below.

\paragraph{Mapping to canonical forms}
We begin by first reading in canonical forms for every informable slot in the MultiWOZ system. For categorical slots, these are defined in a schema file, as released with MultiWOZ 2.1 \cite{eric_multiwoz_2020}. For non-categorical slots, we read in values from the database defined with the original MultiWOZ data collection \cite{budzianowski2018large}. Neither source of information contains dialogue data, only information defining the task. The taxi and train service have informable slots for departure and destination locations. In addition to the locations listed for these slots in a database (i.e. scheduled train journeys), we accept the name of any entity which has an address as a canonical form for these slots. For time slots we consider any time represented in "hh:mm" form as canonical. Overall, this gives us a mapping from a slot name $s_i$ to a set of canonical forms $\mathcal{C_i}$ for all slot names. 

Given a slot name $s_i$ and a slot value surface form $v_j$, we select the correct canonical form $c_j$ as follows: (1) we first generate a set of aliases for $v_j$. These are acceptable re-phrasings of $v_j$, such as adding the leading article "the", a domain specifying suffix such as "hotel" or "museum", or switching numbers to/from digit form (e.g. "one" $\leftrightarrow$ "1"). We then consider a surface form $v_j$ as mapped to a canonical form $c_j$ if any of the aliases $a_j \in A_j$ is a fuzzy match for the canonical form $c_j$, using the \texttt{fuzz.ratio} scorer in the \texttt{fuzzywuzzy} \footnote{\url{https://pypi.org/project/fuzzywuzzy/}} package. We require a score of 90 or higher, and verify in the development data that no surface form maps to more than one canonical form. 

\paragraph{Choosing the most likely surface form} While in a real world dialogue system we would only need to link to canonical forms, \textbf{gold dialogue state states in MultiWOZ are themselves annotated with surface forms}, not always matching the name of the entity in the database and occasionally disagreeing on an entity name. So as to not alter the evaluation process and make sure we can fairly compare to prior work, we use the training data available in each experimental setting to choose the most likely surface form for a given canonical form $c_j$. To do this, we simply count the occurrences of each surface form in the gold labels of the training set for that experiment, and select the most frequently occurring one for $c_j$. However for low data regimes, we often do not observe all canonical forms. Following numerous prior works, we make use of the ontology file released with the dataset \cite{eric_multiwoz_2020, ye_multiwoz_2022}, which lists all observed surface forms for a slot name, and treat each of these as if we had seen them 10 times. This serves as a smoothing factor for selecting the most likely surface form. For the zero-shot experiments, we use only the counts derived from the ontology file, as we have no training data to observe.

Overall, we find this approach to normalization to be robust when compared to other works, which rely on hard-coded fixes for commonly observed typos. Further, our normalization can be initialized with any similarly formatted system definition and data set, allowing for use in other domains. 

To verify that our approach to normalization is not the key factor distinguishing our performance from previous methods, we apply it to a faithful re-implementation of our IC-DST Codex baseline \cite{hu_-context_2022} in our ablation in \autoref{tab:ablation}.

\section{Prompt Examples}

Please see our GitHub repository for prompt examples: \href{https://github.com/jlab-nlp/RefPyDST}{https://github.com/jlab-nlp/RefPyDST}.

\section{Implementation Details}
\label{sec:appendix-impl-details}
\subsection{Hyperparameters}
All hyperparameter tuning is performed using a 10\% split of the development set (100 dialogues) and manual tuning. We find that a smaller choice for $p$ (0.7) in nucleus sampling helps performance in the zero-shot setting. Similarly, we find that in order to select a diverse set of examples, we need to scale $\alpha$. We use $\alpha=0.2$ for the 1\% \& 5\% settings, $\alpha=0.3$ for 10\%, and $\alpha=0.5$ for the full setting. For the full setting, we also increase the the number of considered examples from the nearest 100 to nearest 200. Across all settings, we compute $PMI^\beta$ with $\beta=0.4$. We use a robust approach to normalizing predicted values (i.e. to resolve mis-spellings, etc.) described in Appendix \ref{sec:appendix-normalization}. We apply this normalization to our strongest baseline (IC-DST Codex) in our ablations (\autoref{sec:results-ablations}).
When computing $P(y|\InvPrompt)$, we clip low token log probabilities at 5e-7 in the few-shot setting and 5e-4 in the zero-shot setting, as the lack of examples leads to poorer calibration in the zero-shot setting.  We also clip full-sequence log probabilities at 1e-7 in the few-shot setting and 1e-5 in the zero-shot setting.

\subsection{Retriever fine-tuning details}
For both our methods and the re-implementation of IC-DST Codex \cite{hu_-context_2022} used in our ablations (\autoref{sec:results-ablations}), we fine-tune the retriever using the \texttt{sentence-transformers} package \cite{reimers-2019-sentence-bert}, following the procedure of \cite{hu_-context_2022}. We begin with pre-trained \texttt{all-mpnet-base-v2} embedding model, which we use as a retriever with nearest neighbors search\footnote{ We use the scipy implementation: \url{https://docs.scipy.org/doc/scipy/reference/generated/scipy.spatial.KDTree.html}}. Each of our retrievers is trained for 15 epochs using the \texttt{OnlineContrastiveLoss}, which computes the contrastive loss proposed by \citet{hadsell_et_all_2006} using only hard positives and hard negatives. For each dialogue turn in the training set, we use $sim_{F_1}$ to define positive and (hard) negative examples as the top and bottom 5\% of the nearest 200 examples, respectively.

\subsection{Arguments to Codex}
For all methods, we make requests to OpenAI Codex with arguments \texttt{engine = 'code-davinci-002'}, \texttt{max\_tokens = 120}, and stop sequences of  either \texttt{['--', '\textbackslash n', ';', '\#']} (IC-DST Codex baseline replication) or \texttt{["\textbackslash n\textbackslash n", "\#", "print("]} (ours). For methods which utilize nucleus sampling \cite{holtzman_curious_2020} with the \texttt{top\_p} parameter. In the few-shot setting, we sample with \texttt{best\_of=10}, keeping only \texttt{n=5} most likely results. In the zero-shot setting, we increase \texttt{best\_of} to 32.

\section{Random Retrieval Ablation}
\label{sec:appendix-random-retrieval}
In \autoref{tab:random-retrieval-ablation}, we compare our retrieval methods to random retrieval, on the 20\% split of the development set used in our previous ablations. For random retrieval, we sample $k$ examples from $D_{train}$ uniformly at random to construct $\Ek$. We find this significantly under-performs our learned retrieval methods, whether selecting the top-$k$ examples or using our diverse decoding approach.

\begin{table}[]
    \centering
    \begin{tabular}{l|r}
        \multicolumn{2}{c}{\textbf{Few-Shot (5\%)}} \\
        \hline
         RefPyDST (random-$k$) & 43.5 \\
         RefPyDST (top-$k$) & 54.6 \\
         \hline
         RefPyDST (full) & \textbf{57.9}  \\
    \end{tabular}
    \caption{MultiWOZ joint-goal accuracy in the 5\% few-shot setting, ablating different retrieval methods. The full model includes both our trained retriever and diverse example decoding methods (\S \ref{sec:methods-retrieval}). Top-$k$ uses the trained retriever but decodes the top-$k$ nearest examples instead of using our diverse decoding procedure. Random retrieval samples $k$ examples from $D_{train}$ uniformly at random}
    \label{tab:random-retrieval-ablation}
\end{table}
\end{document}